# Mass Classification Method in Mammogram Using Fuzzy K-Nearest Neighbour Equality

*Abstract:* Mass classification of objects is an important area of research and application in a variety of fields. In this paper, we present an efficient computer-aided mass classification method in digitized mammograms using Fuzzy K-Nearest Neighbour Equality (FK-NNE), which performs benign or malignant classification on region of interest that contains mass. One of the major mammographic characteristics for mass classification is texture. FK-NNE exploits this important factor to classify the mass into benign or malignant. The statistical textural features used in characterizing the masses are Haralick and Run-length features. The main aim of the method is to increase the effectiveness and efficiency of the classification process in an objective manner to reduce the numbers of false-positive of malignancies. In this paper proposes a novel Fuzzy K-Nearest Neighbour Equality algorithm for classifying the marked regions into benign and malignant and 94.46% sensitivity, 96.81% specificity and 96.52% accuracy is achieved that is very much promising compare to the radiologist's accuracy.

## I. Introduction

Breast cancer is the most common of all cancers and is the leading cause of cancer deaths in women worldwide, accounting for more than 1.6% of deaths and case fatality rates are highest in low-resource countries. In India the average age of the high risk group in India is 43-46 years unlike in the west where women aged 53-57 years are more prone to breast cancer. As there is no effective method for its prevention, the diagnosis of breast cancer in its early stage of development has become very crucial for the prevention of cancer. Computer-aided diagnosis (CAD) systems play an important role in earlier diagnosis of breast cancer. Classifiers play an important role in the implementation of intelligent system to identify the breast cancer from mammogram. The features are given as input to the classifiers to classify microcalcifications into benign and malignant. The Back Propagation Neural network is tested by the Jack Knife method [1]. Hassanien and Ali presented an enhanced rough set approach for attribute reduction and generating classification rules from digital mammogram. The classifier model was built and quadratic distances similarly; function is used for matching process.

This Paper is organized as follows. Section II presents related work. Section III describes pre-processing work using new filtering techniques, new segmentation techniques and features extraction techniques. The flow diagram represents the step of the processing. After features extraction and selection, how the feature to be classify the technique that is given in Section IV. Section V presents experimental results. Finally, Section VI presents conclusion.

## II. Related Work

It allows partial membership of an object to different classes, and also takes into account the relative importance (closeness) of each neighbour with respect to the test instance. However, as Sarkar correctly argued in [2], the Fuzzy Neural Network (FNN) algorithm has problems dealing adequately with insufficient knowledge. Cornelis, C., et al. [3] introduced vague quantifiers like "some" or "most" into the definition of upper and lower approximation. Coenen, F., [4] proposed CPAR (Classification based on Predictive Association Rules) algorithm is an extension of PRM (Predictive Rule Mining) which in turn is an extension of FOIL (First Order Inductive Learner) algorithm. Mahmoud, R., et al. [5] proposed approach is performed in two stages. In the first stage, the system separates segments of the image that may correspond to tumors using a combination of morphological operations and a region growing technique. In the second stage, segmented regions are classified as normal, benign, or malignant tissues based on different measurements.

Cheng, H.D., et al. [6] discussed microcalcifications and masses are the two most important indicators of malignancy, and their automated detection is very valuable for early breast cancer diagnosis. Noel pérez et al. [7] described a novel CAD tool that combines digital image processing and artificial neural networks among others techniques to diagnose mammography Pathological Lesions (PL) (as benign or malignant tissues) on GRID environments. Erkang Cheng et al. [8] proposed using normalized Histogram Intersection (HI) as a similarity measure with the K-nearest neighbour (K-NN) classifier. Furthermore, by taking advantage of the fact that HI forms a Mercer kernel, HI is combined with Support Vector Machines (SVM), which further improves the classification performance. A hybrid method of data mining technique is used to predict the texture features which play a vital role in classification.

### III. Pre-Processing Work

Three Hundred and Thirty Two of mammograms are obtained from the MIAS database (ftp://peipa.essex.ac.uk) to analyze the proposed methods. In this chapter, New Filter-I and New Filter-II have been proposed and applied to enhance the mammogram images. The mammogram images are normalized using Max-Min method. The hybridization of different methods based Fuzzy approaches have been proposed for pectoral muscle region. The next work is breast border and edge detection using hybridization of Fuzzy, Canny and Gradient Edge algorithms. Using the border points as references, the mammogram images are aligned and subtracted to extract the suspicious region and background. The two methods of breast region segmentation used to mammogram images to extract the suspicious regions. In the case of pairs of images, the Fuzzy entropy and weighted Fuzzy entropy based on multi-thresholding is used to extract the suspicious region from the digital mammograms. In the last work, microcalcification (Region of Interest (ROI)) process is applied with Modified Ant Colony Optimization and Modified Water Shed Transform algorithms.

The Haralick features from the textural description methods such as Surrounding Region Dependency Matrix, Spatial Grey Level Dependency Matrix, and Grey Level Difference Matrix and run length features from the texture description method Grey Level Run-Length Matrix are extracted from segmented mammogram images for further analysis. The reduced features are selected using two kinds of tolerances rough set based quick reduct, relative reduct and unsupervised Swarm Particle Optimization relative reduct algorithms from the features extracted.

### IV. Proposed Work

However, in many pattern recognition problems, the classification of an input pattern is based on data where the respective sample sizes of each class are small and possibly not representative of the actual probability distributions, even if they are known. In these cases, many techniques rely on some notion of similarity or distance in feature space, for instance, clustering and discriminant analysis. One of the problems encountered in using the K-NN classifier is that normally each of the sample vectors is considered equally important in the assignment of the class label to the input vector. This frequently causes difficulty in those places where the sample sets overlap. The typical vectors are given as much weight as those that are truly representative of the clusters. Another difficulty is that once an input vector is assigned to a class, there is no indication of its "strength" of membership in that class

FK-NN uses concepts from fuzzy logic to assign degree of membership to different classes while considering the distance of its K-Nearest Neighbours. Points closer to the query point contributes larger value to be assigned to the membership function of their corresponding class in comparison to far way neighbours. Class with the highest membership function value is taken as the winner. Since that time researchers have found numerous ways to utilize this theory to generalize existing techniques and to develop new algorithms in pattern recognition and decision analysis.

*A.     Fuzzy K-Nearest Neighbour Equality (FK-NNE) Algorithm*

This section proposes a novel Fuzzy K-Nearest Neighbour Equality algorithm. It assigns class membership to a sample vector rather than assigning the vector to a particular class. The advantage is that no arbitrary assignments are made by the algorithm. The basis of the algorithm is to assign membership as a function of the vector's distance from its K-Nearest Neighbours and those neighbours' memberships in the possible classes. The Fuzzy algorithm is similar to the crisp version in the sense that it must also search the labeled sample set for the K-Nearest Neighbours. Beyond obtaining these K samples, the procedures differ considerably. The mean distance was calculated from K-points to testing data. The minimized mean distance based the output class values are stored.

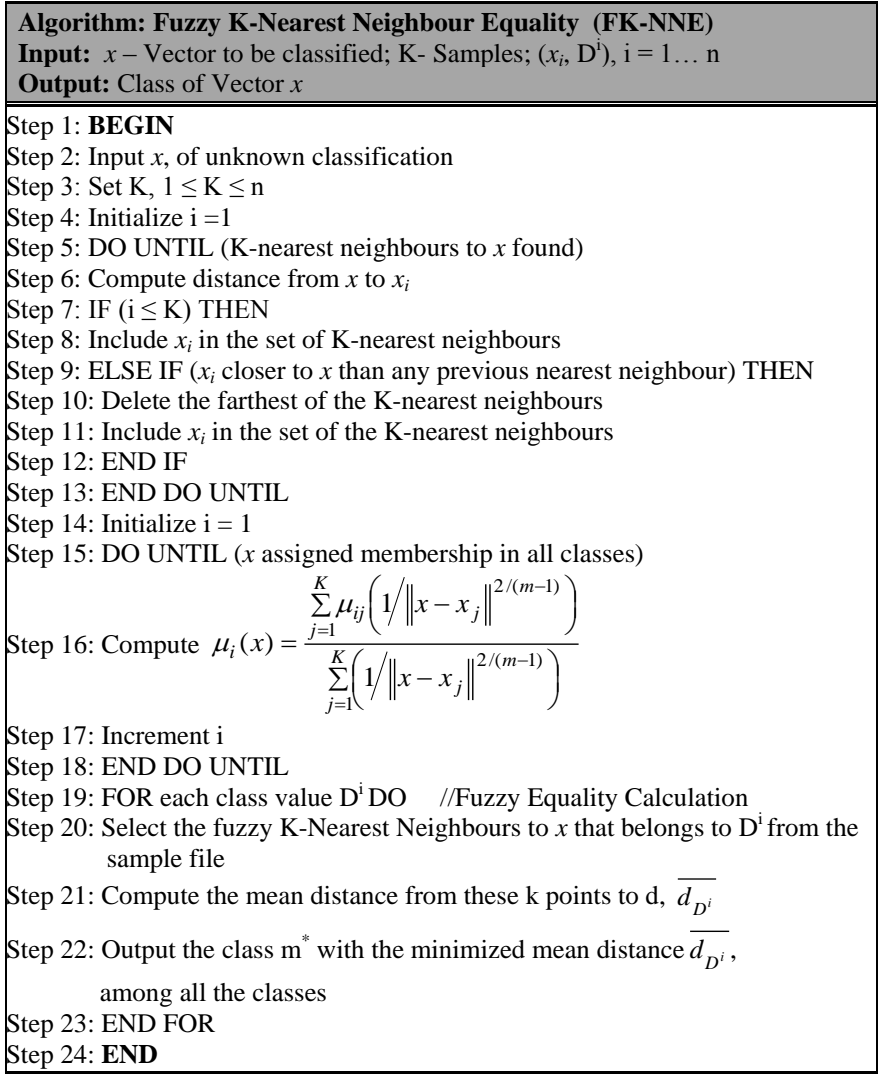

**Figure 1: The FK-NNE Algorithm**

Compared with K-NNE, FK-NNE can achieve acceptable accuracy rate. An improvement over the K-NNE classifier is the Fuzzy K-NN Equality classifier, which uses concepts from fuzzy logic to assign degree of membership to different classes while considering the mean distance of its K-Nearest Neighbours. Figure 1 shows that the Fuzzy K-Nearest Neighbourhood Equality algorithm.

### V. Experimental Results

**Table I Average classification results of classification algorithms**

| Methods | Sensitivity | Specificity | Accuracy |
|---|---|---|---|
| K-NN | 0.8986 | 0.8962 | 0.9125 |
| K-NNE | 0.9311 | 0.9436 | 0.9534 |
| FK-NN | 0.9084 | 0.9222 | 0.9342 |
| FK-NNE | 0.9446 | 0.9681 | 0.9652 |

The results tabulated in Table I, clearly show that the different classification models discriminate malignant and benign with different accuracy. And also it also shows tht the classification accuracy achieved using FK-NNE is much better than others. It is observed that the maximum and minimum classification accuracies are 98% and 91% with FK-NNE and K-NN classifier respectively shown in figures 2 and 3.

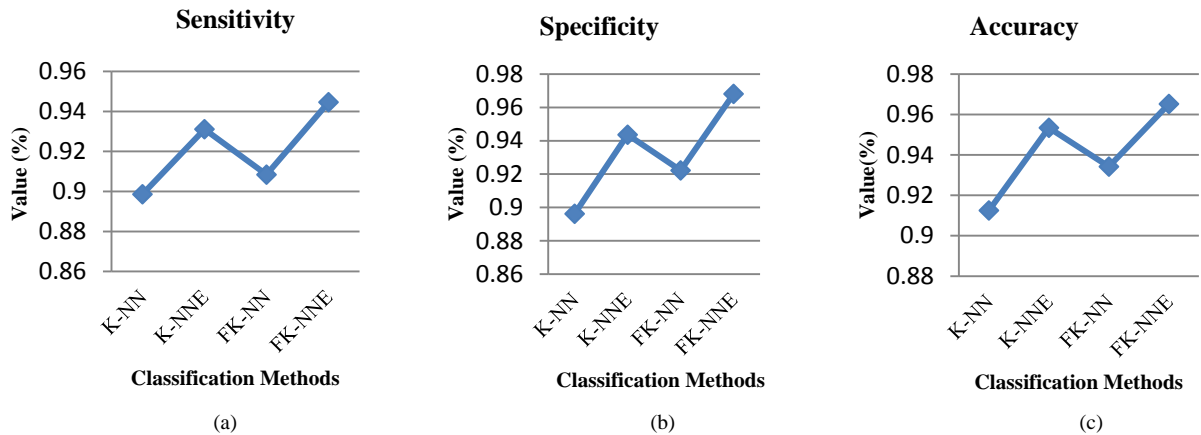

Figure 2: Performance of Sensitivity, Specificity and Accuracy for K-NN, F-KNN, K-NNE and FK-NNE Classifiers

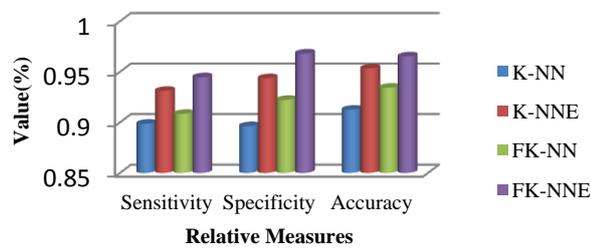

Figure 3: Relative Performance measures for K-NN, FK-NN, K-NNE and FK-NNE and Classifiers

The selected features discriminate between malignant masses and benign masses on mammogram images with 91% accuracy, 90% sensitivity and 90% specificity levels that are relatively poorer when compared to others. FK-NN yielded an accuracy of 93% for distinguishing malignant and benign masses on mammogram images. It is 2% higher than K-NN. The K-NNE classifier achieved an accuracy of 95% where 93% sensitivity and 94% specificity. It is 1% higher than FK-NN. The FK-NNE classifier achieved an accuracy of 97% where 94% sensitivity and 97% specificity. It is 2% higher than K-NNE. ROC curve is generated for the results of four classifiers and exposed in the figure 4.

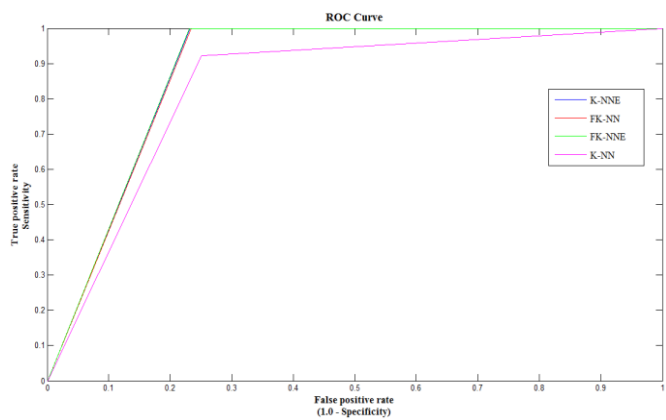

Figure 4: ROC Curves for K-NN, FK-NN, K-NNE and FK-NNE Classifiers

Area under the ROC curve is an important criterion for classifier. The most popular summary measure of accuracy is the area under the ROC curve, often denoted as area under curve. The area under the ROC curve represents the probability of a random positive sample to receive a better score than a random negative sample. The value of area under curve ranges from 0.5 to 1.0 that indicates chance to perfect discrimination. The diagnostic test is more accurate when area is larger. The computed value of area under curve fo each classifier is recorded in Table II.

**Table II: Performance of Classification Algorithms using Area under Curve**

| Algorithms | $A_Z$ Value |
|---|---|
| K-NN | 0.9125 |
| K-NNE | 0.9634 |
| FK-NN | 0.9452 |
| FK-NNE | 0.9734 |

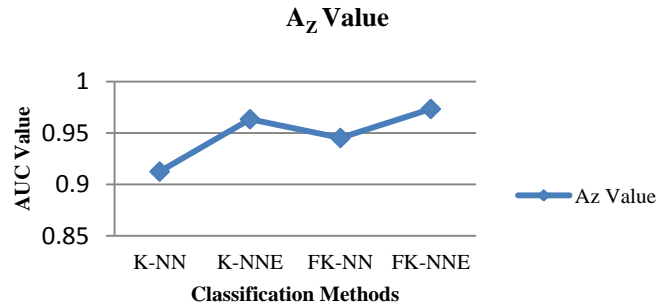

**Figure 5: Areas ($A_z$) under ROC curves for the classifier of K-NN, FK-NN, K-NNE and FK-NNE**

From the Table II, the best area under curve value is 0.9734 for FK-NNE followed by K-NNE, FK-NN and K-NN are 0.9634, 0.9452 and 0.9125 respectively. Figure 5 represents the areas under ROC curves for the proposed classifiers.

## VI. Conclusion

The early detection of breast cancer from the mammogram images is one of the most challenging tasks. Accuracy is also most important in the field of medical diagnosis using images. The performance of four classifiers namely FK-NNE, K-NNE, FK-NN and K-NN have been constructed, and these are investigated for the task of breast cancer classification using mammogram images. The proposed FK-NNE classification accuracy is higher when compared to existing K-NNE, FK-NN and K-NN classifiers. The experimental result reveals that the FK-NNE classifier achieves better classification accuracy than others.

## VIII. Acknowledgments

The First author would like to thank the University Grant Commission, India (F. No. 41-1361/2012(SR)) for his research.